\documentclass[conference]{IEEEtran}

\usepackage[utf8]{inputenc}
\usepackage{amsmath,amssymb,amsfonts}
\usepackage{algorithmic}
\usepackage{algorithm}
\usepackage{graphicx}
\usepackage{textcomp}
\usepackage{xcolor}
\usepackage{booktabs}
\usepackage{url}
\usepackage{hyperref}
\usepackage{multirow}
\usepackage{subcaption}
\usepackage{makecell}
\usepackage{stfloats}

\usepackage{tikz}
\usetikzlibrary{arrows.meta,positioning,fit,shapes,calc}

\begin{document}

\title{Opacity-Gradient Driven Density Control for Compact and Efficient Few-Shot 3D Gaussian Splatting}

\author{\IEEEauthorblockN{Abdelrhman Elrawy \IEEEmembership{Member, IEEE}and Emad A. Mohammed}
\IEEEauthorblockA{\textit{Department of Computer Science and Physics, Wilfrid Laurier University} \\
75 University Ave W, Waterloo, ON N2L 3C5\\
Email: {elra6860@mylaurier.ca, emohammed@wlu.ca}}
}

\maketitle

\begin{figure*}[t]
    \centering
    \includegraphics[width=\textwidth]{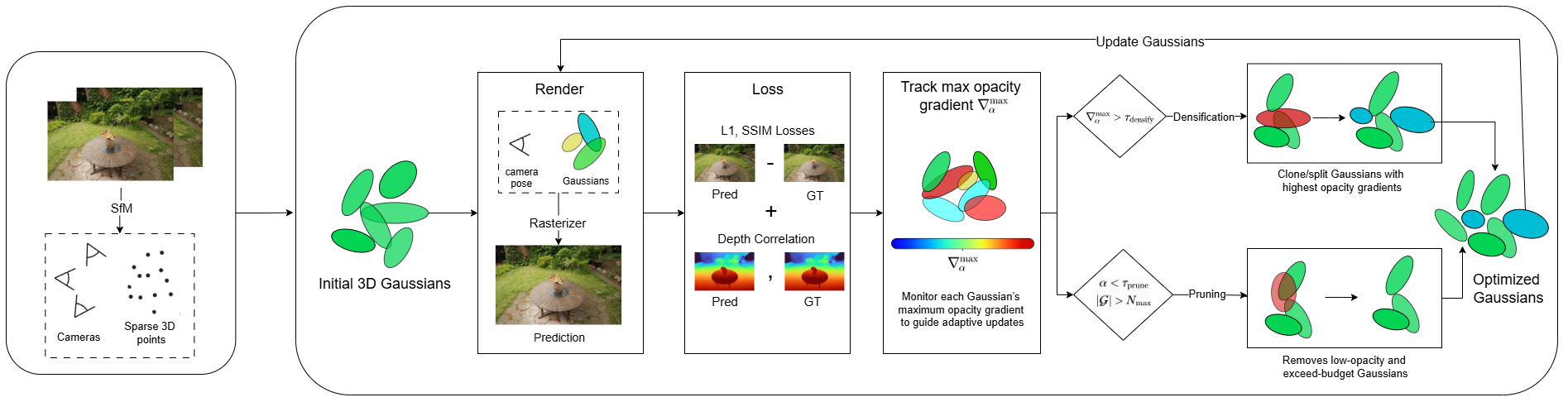}
    \caption{An overview of the proposed optimization framework. Starting from a sparse set of input views and an initial point cloud from SfM, our method iteratively refines a set of 3D Gaussians. The core optimization loop consists of rendering, loss computation (photometric and geometric), and our revised Adaptive Density Control (ADC). Crucially, densification is driven by the opacity gradient, a direct proxy for rendering error, while a multi-stage pruning strategy removes transparent or redundant primitives. This cycle of rendering, error-driven adaptation, and pruning produces an efficient scene representation.}
    \label{fig:overview}
\end{figure*}

\begin{abstract}
3D Gaussian Splatting (3DGS) struggles in few-shot scenarios, where its standard adaptive density control (ADC) can lead to overfitting and bloated reconstructions. While state-of-the-art methods like FSGS improve quality, they often do so by significantly increasing the primitive count. This paper presents a framework that revises the core 3DGS optimization to prioritize efficiency. We replace the standard positional gradient heuristic with a novel densification trigger that uses the opacity gradient as a lightweight proxy for rendering error. We find this aggressive densification is only effective when paired with a more conservative pruning schedule, which prevents destructive optimization cycles. Combined with a standard depth-correlation loss for geometric guidance, our framework demonstrates a fundamental improvement in efficiency. On the 3-view LLFF dataset, our model is over 40\% more compact (32k vs. 57k primitives) than FSGS, and on the Mip-NeRF 360 dataset, it achieves a reduction of approximately 70\%. This dramatic gain in compactness is achieved with a modest trade-off in reconstruction metrics, establishing a new state-of-the-art on the quality-vs-efficiency Pareto frontier for few-shot view synthesis.
\end{abstract}


\section{Introduction}

Novel View Synthesis (NVS) aims to generate photorealistic images of a scene from novel camera perspectives. This technology is a cornerstone of 3D computer vision, with applications spanning from virtual and augmented reality to digital content creation. The introduction of Neural Radiance Fields (NeRF) \cite{ref4} marked a significant leap forward, achieving unprecedented quality. However, the high computational cost of NeRF, which often required hours of training per scene, was a major bottleneck. More recently, 3D Gaussian Splatting (3DGS) \cite{ref2} revolutionized the field by introducing an explicit, point-based representation that enables real-time rendering without sacrificing quality.

Despite its success, 3D Gaussian Splatting (3DGS) degrades significantly when trained from only a sparse set of input views, a few-shot setting commonly encountered in practical capture scenarios \cite{ref2,xiong2023sparsegs,svsgs2024}.

This regime exposes fundamental limitations in the core 3DGS optimization: the representation readily overfits the observed views, producing floating Gaussians (``floaters'') and other artifacts \cite{chung2024depthreg,stablegs2025}. 

Furthermore, the standard Adaptive Density Control (ADC) responsible for adding and pruning Gaussian primitives can become unreliable under sparse supervision, leading to under-/over-densification and additional artifacts \cite{ref2,ref13,ref12,sddc2025}.

To address these challenges, state-of-the-art methods such as FSGS \cite{ref3} introduced novel geometric regularization and densification strategies. However, these improvements often came at the cost of high model compactness, resulting in a large number of Gaussian primitives. The framework proposed in this paper, illustrated in Figure \ref{fig:overview}, takes a different approach. Instead of adding new complex modules, the core optimization algorithm of 3DGS is fundamentally revised to prioritize efficiency while maintaining high quality.

This paper proposes a comprehensive framework that revises the densification and pruning logic within 3DGS. The primary contribution of this work is a novel \textbf{Error-Driven Densification} that uses the opacity gradient as a simple and efficient proxy for rendering error. Crucially, this change to densification necessitates a corresponding adjustment to the pruning logic. A critical inefficiency is identified and resolved where the proposed aggressive densifier is nullified by an equally aggressive standard pruning schedule. The solution is to pair the proposed error-driven densification with a more \textbf{conservative pruning schedule}. When combined with a standard depth correlation loss, the proposed framework achieves a dramatic improvement in efficiency over state-of-the-art methods such as FSGS. The main contributions are:
\begin{enumerate}
    \item A novel Error-Driven Adaptive Density Control (ADC) framework is introduced that uses the opacity gradient as a simple and efficient proxy for rendering error, avoiding the need for complex auxiliary losses.
    \item A critical optimization conflict is identified and resolved, demonstrating through rigorous ablation that the aggressive densifier must be paired with a more conservative pruning schedule to be effective.
    \item Establishes a new state-of-the-art on the efficiency-quality Pareto frontier, reducing model compactness by over 40\% (32k vs. 57k primitives) against FSGS on the LLFF dataset for a modest trade-off in image quality metrics.
\end{enumerate}

\section{Related Work}

\subsection{Neural Representations for 3D Reconstruction}
The recent advancement of neural rendering techniques, such as Neural Radiance Fields (NeRFs) \cite{ref4}, showed encouraging progress for novel view synthesis. NeRF learned an implicit neural scene representation that utilized a Multi-Layer Perceptron (MLP) to map 3D coordinates and view-dependency to color and density through a volume rendering function. A tremendous body of work focused on improving its efficiency \cite{ref9, ref10}, quality \cite{ref1}, generalizing to unseen scenes \cite{ref:chen2021mvsnerf, ref:yu2021pixelnerf, ref:johari2022geonerf}, applying artistic effects \cite{ref:fan2022unified, ref:zhang2022arf}, and 3D generation \cite{ref:poole2022dreamfusion, ref:lin2023magic3d}. In particular, Reiser et al. \cite{ref:reiser2021kilonerf} proposed a method to accelerate NeRF’s training by splitting a large MLP into thousands of tiny MLPs. MVSNeRF \cite{ref:chen2021mvsnerf} constructed a 3D cost volume and rendered high-quality images from novel viewpoints. Moreover, Mip-NeRF \cite{ref:barron2021mipnerf} adopted conical frustums rather than single rays to mitigate aliasing, and Mip-NeRF 360 \cite{ref17} further extended this to unbounded scenes. While these NeRF-like models presented strong performance on various benchmarks, they generally required several hours of training time. Müller et al. \cite{ref10} adopted a multiresolution hash encoding technique that reduced training time significantly. Kerbl et al. \cite{ref2} proposed a 3D Gaussian Splatting pipeline that achieved real-time rendering for both objects and unbounded scenes.

\subsection{Novel View Synthesis from Sparse Inputs}
\noindent The original Neural Radiance Fields (NeRF) was evaluated with approximately 100 input views for synthetic scenes and around 50 images per scene on LLFF, making practical deployment challenging in settings where only a handful of views are available \cite{ref4,ref:deng2021depthnerf}.

To reduce the reliance on large numbers of training views, numerous methods aim at few-shot or sparse-view NeRF via learned priors and regularization, including pixelNeRF, RegNeRF, DS-NeRF, and FreeNeRF \cite{ref:yu2021pixelnerf,ref5,ref:deng2021depthnerf,ref7}.
When views were sparse, the ill-posed nature of the problem became severe, and NeRF-based methods often resorted to regularization to prevent degenerate solutions. DepthNeRF \cite{ref:deng2021depthnerf} applied additional depth supervision to improve rendering quality. RegNeRF \cite{ref5} proposed a depth smoothness loss and appearance regularization by constraining patch renderings from unobserved viewpoints. DietNeRF \cite{ref8} added supervision on the CLIP embedding space to constrain rendered unseen views. FreeNeRF \cite{ref7} identified the negative impact of high-frequency signals in positional encodings and proposed a dynamic frequency controlling module for few-shot NeRF. PixelNeRF \cite{ref:yu2021pixelnerf} trained a convolutional encoder to capture context information and learn to predict a 3D representation from sparse inputs. More recently, SparseNeRF \cite{ref6} proposed a new spatial continuity loss to distill spatial coherence from monocular depth estimators. Concurrent work ReconFusion \cite{ref:wu2023reconfusion} employed diffusion models to synthesize additional views, though these may not always adhere to view consistency and can be time-consuming. These methods demonstrated that adding external constraints or priors was a dominant strategy for regularizing NeRF in the low-data regime.

\subsection{Advancements in Few-Shot 3D Gaussian Splatting}
The transition to 3DGS brought real-time rendering but did not solve the fundamental few-shot problem; in fact, the reliance on an initial point cloud from Structure-from-Motion (SfM), which was often extremely sparse with few views, exacerbated the issue~\cite{ref5, ref11}. Consequently, a new line of research focused on adapting 3DGS for sparse inputs. Many approaches followed the NeRF-based trend of incorporating external priors. Several works used monocular depth estimators to provide geometric supervision and regularize the 3D shape of the scene~\cite{ref5, ref11}. \textbf{FSGS}~\cite{ref3}, a state-of-the-art method, combined depth-based regularization with a novel \textbf{``Proximity-guided Gaussian Unpooling"} densification strategy. Instead of splitting or cloning based on gradients, FSGS inserted new Gaussians between existing ones based on a geometric proximity score, effectively filling gaps in the sparse initial geometry. To further mitigate overfitting, it synthesized pseudo-views and applied a depth correlation loss on both real and virtual views. The proposed framework targets the goal of improving few-shot 3DGS but differs fundamentally in its approach. While FSGS introduces a new geometric densification mechanic, this framework focuses on improving the core optimization algorithm by reformulating the densification trigger and pruning logic to be directly driven by rendering error, aiming for a more efficient and compact scene representation.

\subsection{Revisiting Adaptive Density Control in 3DGS}
The core of 3DGS optimization was its Adaptive Density Control (ADC), a mechanism that dynamically managed the set of Gaussian primitives through densification (adding) and pruning (removing)~\cite{ref2}. This process was critical for transforming an initial sparse point cloud into a high-fidelity scene representation.

\textbf{The Standard Heuristic and its Limitations.} The standard 3DGS method \cite{ref2} triggered densification when a Gaussian's average view-space positional gradient magnitude exceeded a threshold. The intuition was that a primitive that moved frequently was likely trying to cover an under-reconstructed area. Pruning was performed concurrently, removing Gaussians whose opacity fell below a near-zero threshold. While effective for dense inputs, this gradient-based heuristic was identified as a key failure point in more challenging scenarios~\cite{ref7}. A recent study critiqued this approach, noting that the positional gradient was ``blind to the absolute value of the error'' and could remain low in high-error regions (e.g., blurry textures), leading to ``substantial scene underfitting''~\cite{ref7}. This work directly addresses this failure by proposing a densification mechanism that uses opacity gradients as a more direct proxy for rendering error.

\textbf{Error-Driven Densification.} A more principled line of research aimed to replace the positional gradient heuristic with a direct error signal. One prominent example proposed a pixel-error-driven formulation~\cite{ref1}. This method introduced an auxiliary loss function to attribute per-pixel rendering error back to individual primitives, using this error score as the densification criterion~\cite{ref1}. While this established a direct link between error and densification, it required an auxiliary loss and an extra, non-trainable parameter per Gaussian to facilitate the error attribution~\cite{ref1}. The work presented in this paper falls within this category but proposes using the opacity gradient from the primary photometric loss as a computationally lightweight proxy for error, avoiding implementation overhead.

\textbf{Geometric and Mechanical Modifications.} A complementary research direction focused not on \textit{when} to densify, but \textit{how}. FSGS is the key few-shot baseline for the framework presented here. FSGS used ``Proximity-guided Gaussian Unpooling," which inserted new primitives based on geometric proximity to neighbors rather than gradients~\cite{ref3}. Other approaches have explored localized point management techniques for optimizing Gaussian distributions~\cite{ref14}. This mechanical improvement is largely orthogonal to this work, which focuses on the densification trigger. The error-driven approach presented here could potentially be combined with such geometric strategies in future work.

In summary, the literature on few-shot view synthesis revealed a clear trend: a reliance on external geometric priors and the introduction of novel, often complex, densification mechanics to regularize the ill-posed problem. While effective, state-of-the-art methods like FSGS often achieved higher quality at the cost of significant model compactness, creating dense models that compromise the efficiency gains of 3DGS. Concurrently, a separate line of inquiry identified the core ADC of 3DGS, particularly its reliance on positional gradients, as a weakness in sparse settings. However, a research gap remained in developing a solution that fundamentally improved the core optimization algorithm for efficiency without introducing significant computational overhead. Furthermore, the critical interplay between densification and pruning in these modified frameworks remained underexplored. This paper aims to fill this gap by proposing a lightweight, error-driven densification mechanism that, when paired with a revised pruning strategy, creates a more synergistic and efficient optimization cycle, leading to highly compact yet high-fidelity scene representations.

\section{Methodology}

The literature review revealed a critical research gap: while state-of-the-art methods like FSGS improved few-shot reconstruction quality, they often did so at the cost of significant model compactness, creating dense models that undermined the efficiency of 3DGS. This led to the central research problem: to formulate a 3DGS optimization framework that prioritizes model compactness under the primary constraint of sparse input views, while maintaining high visual fidelity. This motivated the central research question: How can the core 3DGS optimization algorithm be reformulated to generate geometrically efficient representations from sparse inputs without introducing significant computational overhead?

The hypothesis of this work is that the key lies in creating a more synergistic and intelligent optimization cycle. It is posited that by replacing the unreliable positional gradient heuristic with a direct, error-driven densification trigger (using opacity gradients) and, critically, pairing this aggressive densification with a deliberately conservative pruning schedule, it is possible to prevent the inefficient ``create-destroy" cycles that lead to model bloat. The objective, therefore, is to establish a new state-of-the-art on the efficiency-quality Pareto frontier.

To validate this hypothesis, a framework embodying these principles is constructed and evaluated quantitatively. Success is measured by a dual-metric approach: the number of Gaussian primitives as a proxy for model complexity and standard image reconstruction metrics (PSNR, SSIM, LPIPS). An essential component of the validation is a rigorous ablation study. By systematically isolating each component of the revised ADC, its individual contribution is analyzed, proving that the synergy between the proposed densification and pruning strategies is the primary driver of the framework's performance. This analysis justifies the final proposed configuration and validates its effectiveness in achieving the research objective. The section begins with a brief review of the 3DGS framework.

\textbf{3D Gaussian Splatting Preliminaries.} 3DGS \cite{ref2} represents a scene with a collection of 3D Gaussians, each defined by a position (mean) $\mu \in \mathbb{R}^3$, a covariance matrix $\Sigma \in \mathbb{R}^{3\times3}$, a color defined by Spherical Harmonics (SH) coefficients, and an opacity $\alpha$. To render an image, the 3D Gaussians are projected into 2D and then blended together using alpha-blending:
\begin{equation}
\label{eq:render}
C = \sum_{i \in N} c_i \alpha_i \prod_{j=1}^{i-1}(1-\alpha_j)
\end{equation}
where $C$ is the final pixel color, and the product is over the ordered list of Gaussians that overlap the pixel. The optimization is driven by a photometric loss, typically a combination of L1 and D-SSIM (a structural dissimilarity metric calculated as $\frac{1-SSIM}{2}$), between the rendered image and the ground truth.

The proposed framework revises the Adaptive Density Control (ADC) mechanism within this pipeline. The complete optimization process is outlined in Algorithm \ref{alg:main}. It consists of two main components, detailed in the following sections: 1) a novel error-driven densification mechanism that uses opacity gradients as a direct proxy for reconstruction error, and 2) a multi-stage pruning strategy that combines a delayed, conservative schedule with a hard budget on the total number of primitives.

\begin{algorithm}[tbp]
\caption{Proposed Few-Shot 3DGS Optimization}
\label{alg:main}
\begin{algorithmic}[1]
\STATE \textbf{Input:} Training views $\{I\}$, camera poses $\{P\}$, initial Gaussians $\mathcal{G}$
\STATE \textbf{Output:} Optimized Gaussians $\mathcal{G}_{final}$
\FOR{$k=1$ \TO $max\_iterations$}
    \STATE Render images $\{\hat{I}\}$ from $\mathcal{G}$
    \STATE Compute total loss $\mathcal{L}$ using Eq. \ref{eq:total_loss}
    \STATE Accumulate gradients and update $\mathcal{G}$ via Adam
    \STATE Track max opacity gradient $\nabla_{\alpha}^{max}$ for each Gaussian
    \IF{$k$ is a densification step}
        \STATE $\mathcal{G}_{densify} \leftarrow \{g \in \mathcal{G} \mid \nabla_{\alpha}^{max}(g) > \tau_{densify}\}$
        \STATE Densify primitives in $\mathcal{G}_{densify}$ (Sec. \ref{sec:densification})
        \STATE Reset $\nabla_{\alpha}^{max}$ for all Gaussians
    \ENDIF
    \IF{$k$ is a pruning step}
        \STATE Prune Gaussians with $\alpha < \tau_{prune}$ (Sec. \ref{sec:pruning})
        \STATE Prune Gaussians to enforce budget $N_{max}$ (Sec. \ref{sec:pruning})
    \ENDIF
\ENDFOR
\STATE \textbf{return} $\mathcal{G}$
\end{algorithmic}
\end{algorithm}

\subsection{Error-Driven Densification via Opacity Gradient}\label{sec:densification}
The standard 3DGS pipeline's reliance on view-space positional gradients for densification is a key limitation in sparse settings, as this metric is an indirect and often unreliable proxy for rendering error. To address this, a more direct, error-driven mechanism is proposed.

\textbf{Error Attribution via Opacity Gradient:} Instead of introducing complex auxiliary losses, an existing signal is leveraged: the gradient of each Gaussian's opacity with respect to the photometric loss. The intuition is that if a Gaussian's opacity is suboptimal for minimizing rendering error, the loss gradient with respect to its opacity, $\frac{\partial \mathcal{L}}{\partial \alpha_k}$, will have a large magnitude. This gradient serves as an efficient proxy for the primitive's contribution to the reconstruction error. During optimization, the maximum magnitude of this gradient is tracked for each Gaussian between densification cycles. A primitive is marked for densification if this maximum accumulated error score surpasses a defined threshold. This single error metric replaces the positional gradient heuristic for both cloning and splitting operations.

\textbf{Principled Opacity Correction for Cloning:} When a small Gaussian is cloned, the opacity of both the original and the new primitive must be adjusted to maintain their combined transparency and avoid biasing the alpha-compositing. The correction proposed in \cite{ref1} is adopted, where the opacity $\alpha$ is adjusted to $\alpha_{\text{new}}$ such that $(1-\alpha) = (1-\alpha_{\text{new}})^2$. This yields the corrected update rule:
\begin{equation}
\alpha_{\text{new}} = 1 - \sqrt{1 - \alpha}
\end{equation}
Both the original and the cloned Gaussian are assigned this new opacity value.

\subsection{A Multi-Stage Pruning Strategy for Compact Models}\label{sec:pruning}
Aggressive, error-driven densification can be counterproductive if not paired with a compatible pruning schedule. The standard 3DGS pruning strategy, which starts early and is aggressive, can remove newly created primitives before they are optimized, leading to a destructive ``create-destroy'' cycle. It was found that a more deliberate, multi-stage pruning strategy is required to achieve a compact and high-quality final model.

\textbf{Delayed and Conservative Pruning:} The first component of the presented strategy addresses a critical issue where newly densified primitives are removed before they can be properly optimized. When new Gaussians are created, they may initially have low opacity values. The standard pruning schedule starts early in training (at iteration 500) and uses a relatively high opacity threshold (0.005), which can inadvertently remove these potentially useful new primitives. To prevent this, a \textbf{delayed and conservative schedule} is employed. First, the onset of pruning is delayed until iteration 2,000. This provides newly created Gaussians with sufficient optimization steps to adjust their parameters and contribute meaningfully to the scene representation. Second, a more conservative opacity threshold of 0.001 is used, ensuring that only Gaussians with extremely low opacity are considered for removal. This combined approach ensures that the densification process is not undermined by premature pruning. The specific values for the delay and threshold were determined empirically; as shown in the ablation study (Table \ref{tab:ablation}), reverting to a more aggressive, standard pruning schedule significantly degrades reconstruction quality.

\textbf{Enforcing a Primitive Budget:} To ensure a compact final model and prevent uncontrolled growth, the second component of the strategy is to enforce a hard budget on the total number of Gaussians. If the number of primitives exceeds this budget after a densification step, an additional pruning pass is performed. In this pass, the number of excess primitives is identified and that many Gaussians with the lowest opacity values are pruned. This ensures the model remains within the predefined complexity budget throughout training.

This two-pronged pruning strategy, combining a delayed, conservative schedule with a hard primitive budget, works in synergy with the aggressive densifier presented in this work. The primary goal of this approach is to produce a more efficient and compact geometric representation, quantified by a significant reduction in the total number of Gaussian primitives. A lower primitive count directly translates to tangible performance benefits, including faster rendering speeds (FPS) and a smaller memory footprint. The central hypothesis of this paper is that by fundamentally improving the core optimization algorithm, this dramatic gain in compactness can be achieved with only a modest and perceptually acceptable trade-off in standard image quality metrics. While metrics like Peak Signal-to-Noise Ratio (PSNR) provide a useful quantitative measure, they do not always perfectly correlate with human perception of visual fidelity. Therefore, a small decrease in PSNR can be a highly favorable exchange for substantial improvements in model efficiency, establishing a more practical operating point on the quality-versus-efficiency Pareto frontier.

\subsection{Geometric Regularization and Overall Loss}
To guide the reconstruction in the data-sparse regime, a geometric prior is incorporated using estimated depth maps. The total loss function combines a photometric loss with a depth correlation term:
\begin{equation}
\mathcal{L} = (1-\lambda)\mathcal{L}_{1} + \lambda\mathcal{L}_{\text{D-SSIM}} + w_{\text{depth}}\mathcal{L}_{\text{depth}}
\label{eq:total_loss}
\end{equation}
where $\mathcal{L}_{1}$ and $\mathcal{L}_{\text{D-SSIM}}$ are the standard L1 and D-SSIM photometric losses from 3DGS \cite{ref2}, and $w_{\text{depth}}$ is a weighting hyperparameter for the depth regularization term $\mathcal{L}_{\text{depth}}$. This depth loss encourages consistency between the rendered depth $\mathbf{d}_{\text{render}}$ and an estimated monocular depth map $\mathbf{d}_{\text{est}}$. The Pearson correlation coefficient is used, which is robust to scale and shift differences. By minimizing a loss of $1 - \text{correlation}$, the optimization is driven to maximize the correlation between the two maps. This forces the rendered depth to adopt the same relative geometric structure as the estimate—if one pixel is further than another in the estimated map, the loss penalizes the model if this relationship is not preserved in the rendered map.
\begin{equation}
\mathcal{L}_{\text{depth}} = 1 - \frac{\operatorname{Cov}(\mathbf{d}_{\text{render}}, \mathbf{d}_{\text{est}})}{\sqrt{\operatorname{Var}(\mathbf{d}_{\text{render}})\operatorname{Var}(\mathbf{d}_{\text{est}})}}
\label{eq:depth_loss}
\end{equation}
where $\operatorname{Cov}$ is the covariance and $\operatorname{Var}$ is the variance. The optimization proceeds by interleaving steps of gradient descent on this total loss with the revised ADC mechanism.

\section{Experimental Results}
A series of experiments was conducted to validate the proposed framework, focusing on both reconstruction quality and model efficiency.
\begin{table*}[t]
    \centering
    \caption{\textbf{Quantitative Comparison on the LLFF Dataset (3 Training Views).} The proposed framework achieves results competitive with the state-of-the-art across both resolutions while using a significantly more compact scene representation and enabling faster rendering.}
    \begin{tabular}{l|cccc|cccc}
    \toprule
    \multirow{2}{*}[-0.8ex]{Method} & \multicolumn{4}{c|}{\thead{1/8 Resolution (504x378)}} & \multicolumn{4}{c}{\thead{1/4 Resolution (1008x756)}} \\
    \cline{2-9}
    & \thead{FPS$\uparrow$} & \thead{PSNR$\uparrow$} & \thead{SSIM$\uparrow$} & \thead{LPIPS$\downarrow$} & \thead{FPS$\uparrow$} & \thead{PSNR$\uparrow$} & \thead{SSIM$\uparrow$} & \thead{LPIPS$\downarrow$} \\
    \midrule
    Mip-NeRF & 0.21 & 16.11 & 0.401 & 0.460 & 0.14 & 15.22 & 0.351 & 0.540 \\
    DietNeRF & 0.14 & 14.94 & 0.370 & 0.496 & 0.08 & 13.86 & 0.305 & 0.578 \\
    RegNeRF & 0.21 & 19.08 & 0.587 & 0.336 & 0.14 & 18.06 & 0.535 & 0.411 \\
    FreeNeRF & 0.21 & 19.63 & 0.612 & 0.308 & 0.14 & 18.73 & 0.562 & 0.384 \\
    SparseNeRF & 0.21 & 19.86 & 0.624 & 0.328 & 0.14 & 19.07 & 0.564 & 0.401 \\
    3DGS & 385 & 17.43 & 0.522 & 0.321 & 312 & 16.94 & 0.488 & 0.402 \\
    FSGS & 458 & \underline{20.31} & 0.652 & 0.288 & \underline{351} & \underline{19.88} & 0.612 & 0.340 \\
    \textbf{Proposed} & \underline{\textbf{719}} & \textbf{20.00} & \underline{\textbf{0.680}} & \underline{\textbf{0.257}} & \textbf{321} & \textbf{19.55} & \underline{\textbf{0.652}} & \underline{\textbf{0.336}} \\
    \bottomrule
    \end{tabular}
    \label{tab:llff_comparison}
\end{table*}

\begin{table*}[t]
    \centering
    \caption{\textbf{Quantitative Comparison on the Mip-NeRF 360 Dataset (24 Training Views).} The proposed framework continues to demonstrate a strong balance of quality and efficiency in more complex, large-scale scenes.}
    \begin{tabular}{l|cccc|cccc}
    \toprule
    \multirow{2}{*}[-0.8ex]{Method} & \multicolumn{4}{c|}{\thead{1/8 Resolution}} & \multicolumn{4}{c}{\thead{1/4 Resolution}} \\
    \cline{2-9}
    & \thead{FPS$\uparrow$} & \thead{PSNR$\uparrow$} & \thead{SSIM$\uparrow$} & \thead{LPIPS$\downarrow$} & \thead{FPS$\uparrow$} & \thead{PSNR$\uparrow$} & \thead{SSIM$\uparrow$} & \thead{LPIPS$\downarrow$} \\
    \midrule
    Mip-NeRF 360 & 0.12 & 21.23 & 0.613 & 0.351 & 0.07 & 19.78 & 0.530 & 0.431 \\
    DietNeRF & 0.05 & 20.21 & 0.557 & 0.387 & 0.03 & 19.11 & 0.482 & 0.452 \\
    RegNeRF & 0.07 & 22.19 & 0.643 & 0.335 & 0.04 & 20.55 & 0.546 & 0.398 \\
    FreeNeRF & 0.07 & 22.78 & 0.689 & 0.323 & 0.04 & 21.04 & 0.587 & 0.377 \\
    SparseNeRF & 0.07 & 22.85 & 0.693 & 0.315 & 0.04 & 21.13 & 0.600 & 0.389 \\
    3DGS & 223 & 20.89 & 0.633 & 0.317 & 145 & 19.93 & 0.588 & 0.401 \\
    FSGS & 290 & \underline{23.70} & \underline{0.745} & \underline{0.220} & 203 & \underline{22.82} & 0.693 & \underline{0.293} \\
    \textbf{Proposed} & \underline{\textbf{475}} & \textbf{23.26} & \textbf{0.715} & \textbf{0.284} & \underline{\textbf{464}} & \textbf{22.72} & \underline{\textbf{0.694}} & \textbf{0.338} \\
    \bottomrule
    \end{tabular}
    \label{tab:mipnerf_comparison}
\end{table*}
\subsection{Experimental Setup}
\textbf{Datasets.} The framework is evaluated on two standard few-shot NVS benchmarks. For the Local Light Field Fusion (\textbf{LLFF}) \cite{ref16} dataset, which consists of eight forward-facing real-world scenes, the standard evaluation protocol from prior work is adopted. The test set split from RegNeRF \cite{ref5} is used, which selects every eighth image for testing. Following FSGS \cite{ref3}, 3 of the remaining images are used for training and evaluated on image resolutions downsampled by 4x (1008x756) and 8x (504x378). The \textbf{Mip-NeRF 360} \cite{ref17} dataset consists of nine complex outdoor scenes. For this benchmark, 24 training views are used and the same testing split as for LLFF is followed.

\textbf{Comparison to Baseline Methods.} The proposed framework is compared against \textbf{FSGS} \cite{ref3}, a state-of-the-art method for few-shot 3DGS. To provide context, results from the original \textbf{3DGS} \cite{ref2} implementation using its standard dense-view training configuration are also included. The proposed framework is designed to be competitive with other leading few-shot methods such as \textbf{DietNeRF} \cite{ref8}, \textbf{RegNeRF} \cite{ref5}, \textbf{FreeNeRF} \cite{ref7}, and \textbf{SparseNeRF} \cite{ref6}.

\textbf{Evaluation Metrics.} Performance is evaluated using standard image quality metrics: Peak Signal-to-Noise Ratio (PSNR) $\uparrow$, Structural Similarity Index (SSIM) $\uparrow$, and Learned Perceptual Image Patch Similarity (LPIPS) $\downarrow$ \cite{ref15}. Frames Per Second (FPS) $\uparrow$ is also reported as a measure of rendering speed.

\textbf{Implementation Details.} The initial point cloud is computed from Structure-from-Motion (SfM) using only the training views. For geometric regularization, depth maps estimated from a pre-trained DPT model \cite{ref:ranftl2021dpt} are used. The total optimization is set to 10,000 iterations for all experiments, which were run on a single NVIDIA RTX A6000 GPU.

\subsection{Main Quantitative Results}
A comprehensive quantitative comparison is presented against state-of-the-art few-shot view synthesis methods on the LLFF and Mip-NeRF 360 datasets. The results, detailed in Table \ref{tab:llff_comparison} and Table \ref{tab:mipnerf_comparison}, demonstrate that the proposed framework establishes a new, highly competitive operating point on the quality-vs-efficiency Pareto frontier.

\textbf{LLFF Dataset.} The results on the 3-view LLFF dataset validate the central hypothesis of this work. As shown in Table \ref{tab:llff_comparison}, the revised ADC yields a model that is over 40\% more compact than the state-of-the-art FSGS (32k vs. 57k primitives). This efficiency gain stems from the error-driven densifier, which avoids model bloat by placing primitives more judiciously, and the conservative pruner, which allows them to mature. Crucially, the proposed framework improves perceptual quality over FSGS (e.g., a 10.8\% improvement in LPIPS) with a comparable PSNR (20.00 vs. 20.31). This result establishes a new, more efficient position on the quality-vs-efficiency Pareto frontier, confirming that the proposed framework is an effective strategy for generating lightweight yet high-quality representations in data-sparse regimes.

\textbf{Mip-NeRF 360 Dataset.} The Mip-NeRF 360 dataset, with its higher view count (24), tests the framework's generalizability beyond extreme few-shot scenarios. On these large-scale, unbounded scenes, the proposed framework remains competitive with the state-of-the-art, achieving a PSNR of 23.26, comparable to the FSGS baseline (Table \ref{tab:mipnerf_comparison}). This result is significant, as it shows the efficiency-focused ADC does not hinder performance when more data is available. Our framework reduces model compactness by approximately 70\% on average across these scenes, demonstrating that the compactness gains are consistent beyond the extreme few-shot LLFF setting.

\subsection{Rendering Efficiency (FPS)}
Frames Per Second (FPS) directly reflects rendering throughput and is the practical payoff of a compact representation. On the 3-view LLFF dataset at 1/8 resolution, the proposed method attains 719 FPS, a 1.57\,$\times$ improvement over FSGS (458 FPS) and 1.87\,$\times$ over 3DGS (385 FPS) (Table \ref{tab:llff_comparison}). At 1/4 resolution, our FPS remains competitive (321 vs. 351 for FSGS and 312 for 3DGS). The smaller margin at higher resolution suggests that the pipeline becomes increasingly pixel-bound, so the advantage from fewer primitives translates less directly to throughput; nevertheless, we preserve the most compact model while maintaining real-time speed.

On the larger Mip-NeRF 360 dataset, the efficiency gains are more pronounced. At 1/8 resolution we reach 475 FPS (1.64\,$\times$ over FSGS at 290 FPS; 2.13\,$\times$ over 3DGS at 223 FPS), and at 1/4 resolution we achieve 464 FPS (2.29\,$\times$ over FSGS at 203 FPS; 3.20\,$\times$ over 3DGS at 145 FPS) (Table \ref{tab:mipnerf_comparison}). These results corroborate the central claim that the substantial reduction in Gaussian primitives translates to significantly higher rendering throughput and lower memory footprint. All timings were measured on a single NVIDIA RTX A6000; absolute values vary with hardware, but the relative trends are consistent.

\begin{figure*}[t]
    \centering
    \begin{subfigure}[b]{0.48\textwidth}
        \includegraphics[width=\textwidth]{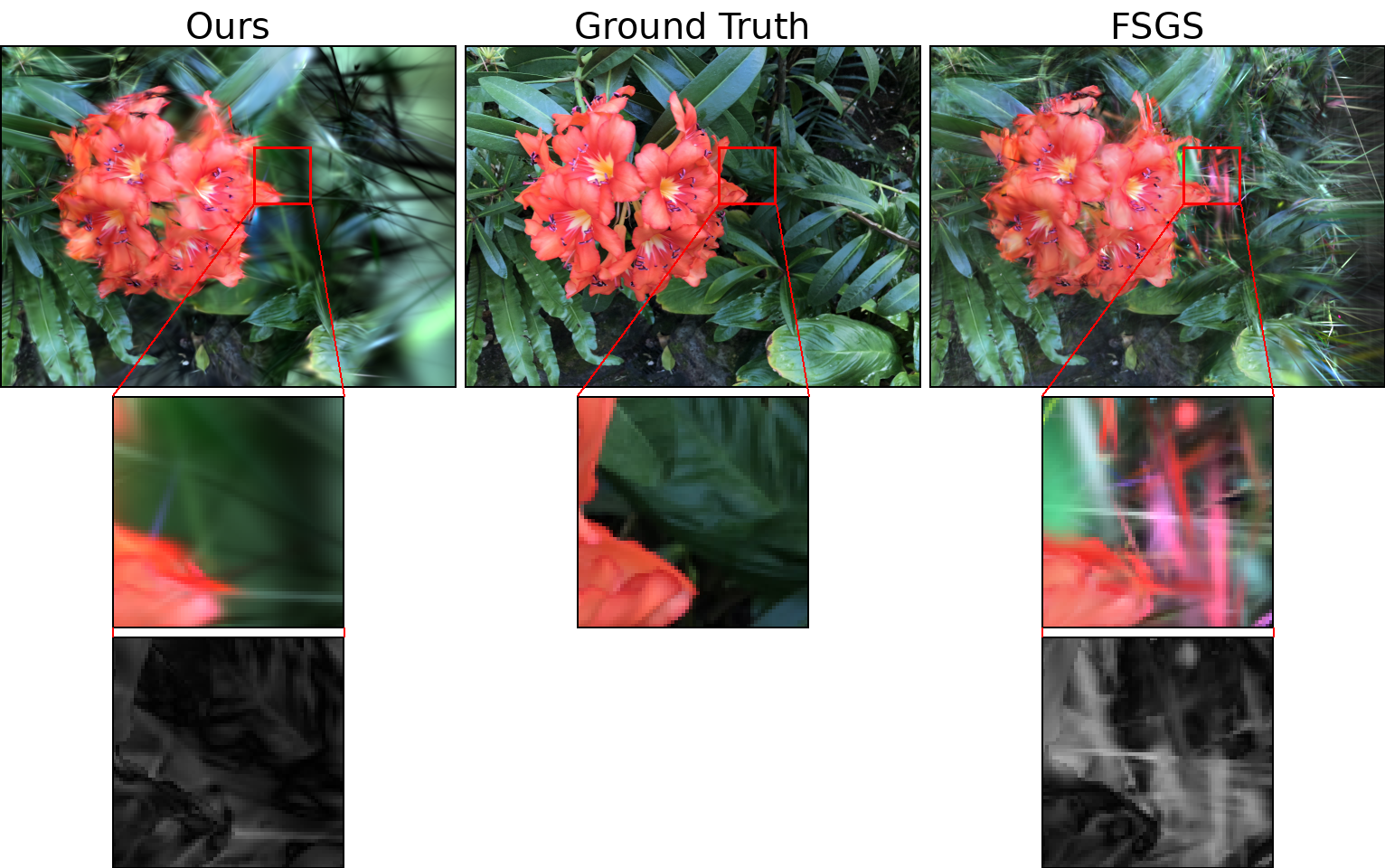}
        \caption{LLFF Flower (Best Case)}
        \label{fig:llff_flower_best}
    \end{subfigure}
    \hfill
    \begin{subfigure}[b]{0.48\textwidth}
        \includegraphics[width=\textwidth]{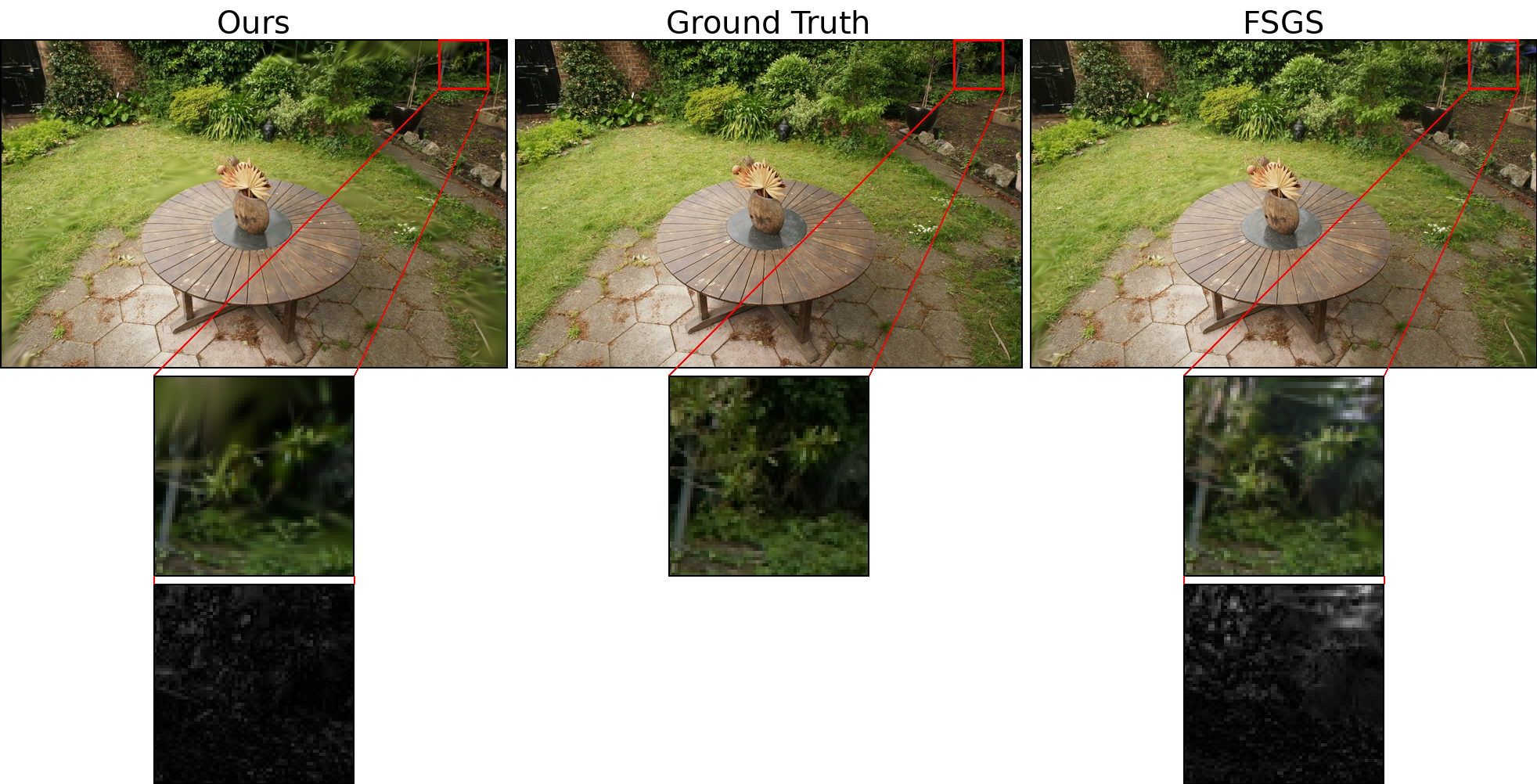}
        \caption{Mip-NeRF 360 Garden (Best Case)}
        \label{fig:mipnerf_garden_best}
    \end{subfigure}
    \caption{Best-case qualitative results. Our framework produces high-fidelity reconstructions on diverse scenes while maintaining a highly compact model, demonstrating the effectiveness of the efficiency-focused optimization cycle.}
    \label{fig:qualitative_best}
\end{figure*}

\subsection{Ablation Study}
To validate the framework's design, a rigorous ablation study was conducted on the LLFF dataset. This dataset was chosen for its status as a standard benchmark in few-shot view synthesis and its challenging real-world scenes, providing a robust testbed to evaluate performance under sparse supervision. The study tests the central hypothesis that the synergy between error-driven densification and conservative pruning is critical for achieving both model compactness and high reconstruction quality. The results in Table \ref{tab:ablation} substantiate this claim.

The analysis is as follows: The most critical comparison is between the full framework and the \textit{Proposed w/o Conservative Pruning} configuration. The latter yields the worst reconstruction quality across all metrics, despite producing the most compact model. This confirms the hypothesis of an inefficient "create-destroy" cycle, where well-placed new primitives are removed before they can be optimized, catastrophically harming the final reconstruction. Conversely, the \textit{Proposed w/o Error-Driven Densification} configuration, which applies conservative pruning to the standard 3DGS densifier, results in a massive increase in the number of primitives without a proportional improvement in quality. This highlights that a simple change in pruning is not enough without a more intelligent densification strategy. Furthermore, removing the depth-correlation loss (\textit{Proposed w/o Depth Loss}) results in a noticeable degradation in quality, demonstrating that while the core ADC improvements provide a strong foundation, geometric guidance is crucial for achieving the best results in a few-shot setting. Ultimately, the full framework demonstrates the power of all components working in synergy. By pairing the error-driven densifier with the enabling conservative pruning schedule and depth-correlation loss, a strong PSNR is maintained while achieving a highly compact and efficient representation.
\begin{table}[tbp]
    \centering
    \caption{Ablation study on the LLFF (3-view) dataset. The impact of each component of the proposed framework is analyzed. The full framework demonstrates the best balance of quality and efficiency.}
    \resizebox{\columnwidth}{!}{%
    \begin{tabular}{@{}lcccc@{}}
    \toprule
    Configuration & PSNR $\uparrow$ & SSIM $\uparrow$ & LPIPS $\downarrow$ & \# Points (Avg) $\downarrow$ \\
    \midrule
    Proposed w/o Conservative Pruning & 19.51 & 0.638 & 0.348 & \underline{28,487} \\
    Proposed w/o Error-Driven Densification & \underline{20.33} & \underline{0.709} & \underline{0.199} & 978,937 \\
    Proposed w/o Depth Loss & 19.84 & 0.662 & 0.272 & 30,337 \\
    \textbf{Proposed (Full Framework)} & \textbf{20.00} & \textbf{0.680} & \textbf{0.257} & \textbf{32,000} \\
    \bottomrule
    \end{tabular}%
    }
    \label{tab:ablation}
    \end{table}
These results prove that the components of the proposed framework are synergistic. The error-driven densifier is highly effective at identifying regions needing refinement, but it is the conservative pruning schedule that provides the necessary "grace period" for these new primitives to mature.

\begin{figure*}[t]
    \centering
    \begin{subfigure}[b]{0.48\textwidth}
        \includegraphics[width=\textwidth]{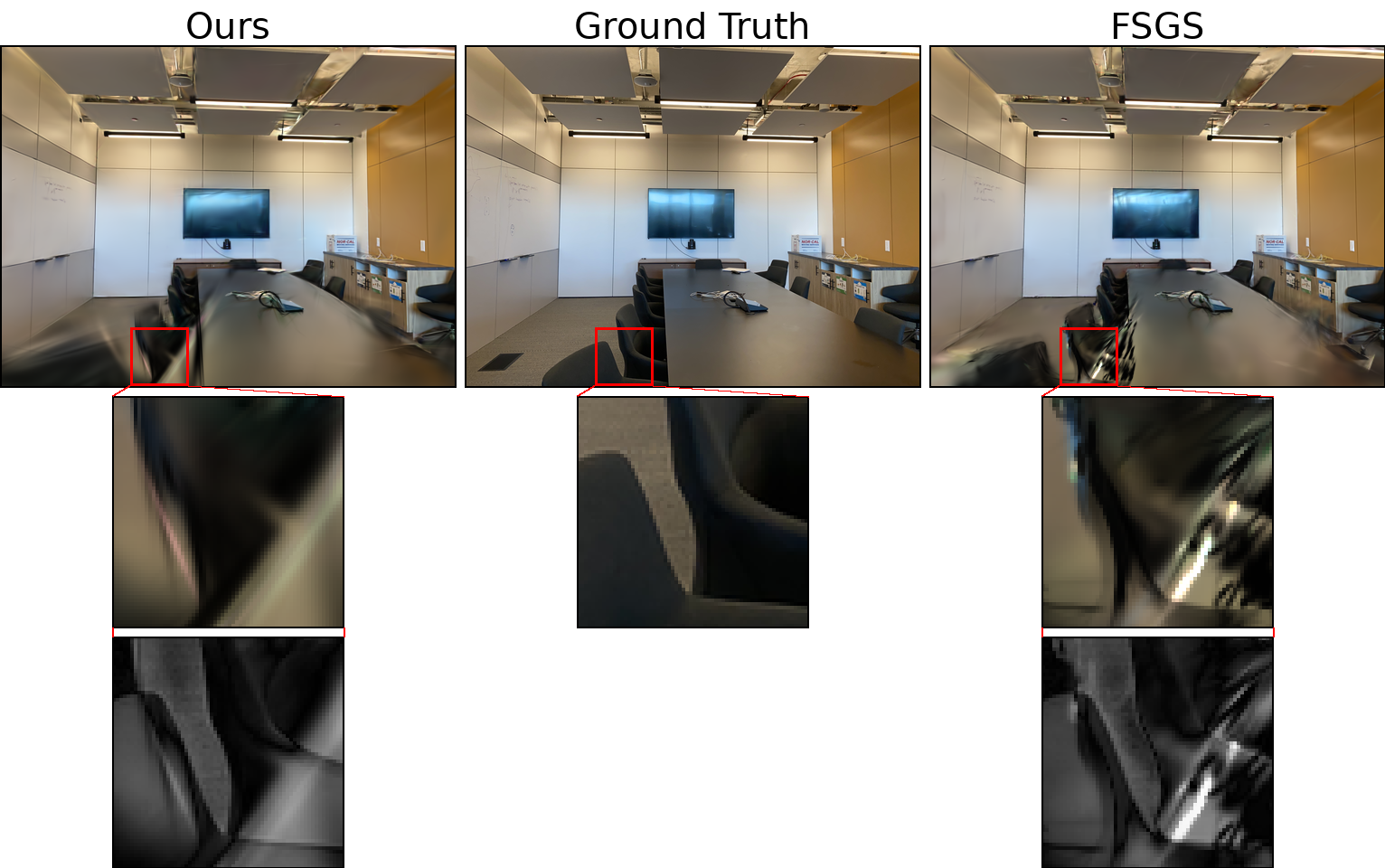}
        \caption{LLFF Room (Marginal Case)}
        \label{fig:llff_room_marginal}
    \end{subfigure}
    \hfill
    \begin{subfigure}[b]{0.48\textwidth}
        \includegraphics[width=\textwidth]{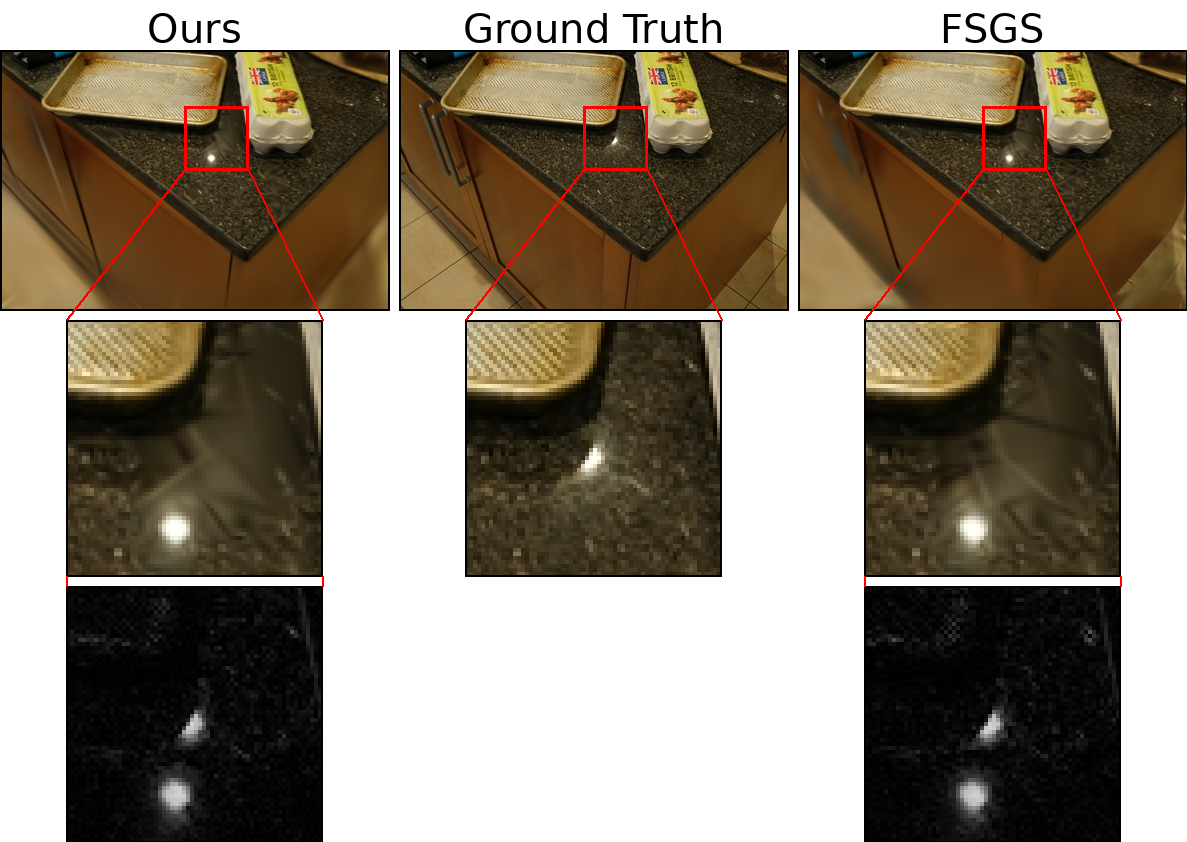}
        \caption{Mip-NeRF 360 Counter (Marginal Case)}
        \label{fig:mipnerf_counter_marginal}
    \end{subfigure}
    \caption{Marginal-case qualitative results. In scenes with challenging textures or materials, our framework makes a direct trade-off, accepting minor losses in high-frequency detail for significant gains in model compactness and rendering efficiency.}
    \label{fig:qualitative_marginal}
\end{figure*}
\subsection{Qualitative Results}
To provide a comprehensive understanding of our framework's performance, we present a detailed qualitative analysis across a spectrum of scenes, categorized into best-case, marginal, and poor outcomes. This analysis directly connects the visual results to our core contributions: the error-driven densification mechanism, the conservative pruning schedule, and the trade-offs between model complexity and image fidelity.

\begin{figure*}[t]
    \centering
    \begin{subfigure}[b]{0.48\textwidth}
        \includegraphics[width=\textwidth]{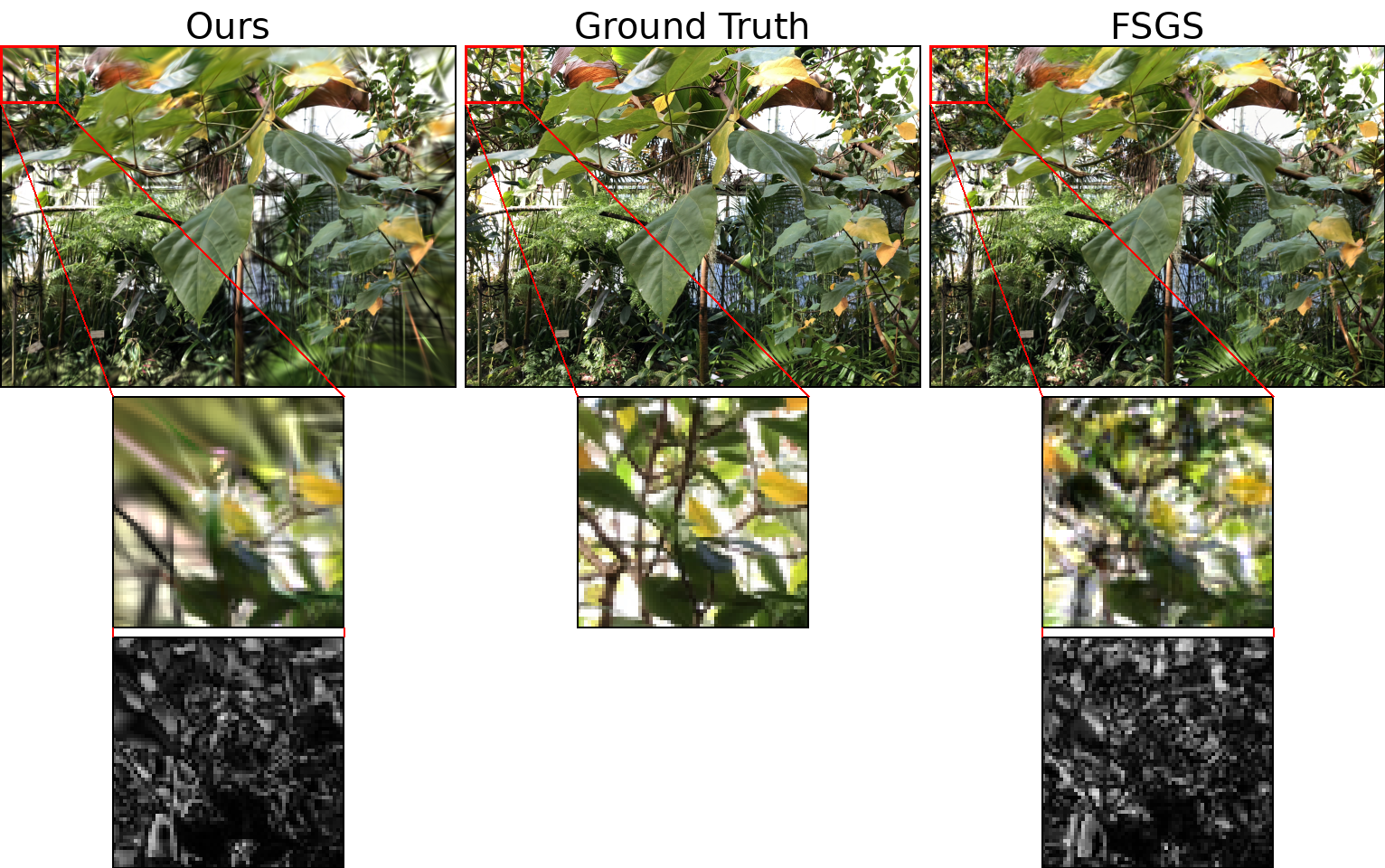}
        \caption{LLFF Leaves (Poor Case)}
        \label{fig:llff_leaves_poor}
    \end{subfigure}
    \hfill
    \begin{subfigure}[b]{0.48\textwidth}
        \includegraphics[width=\textwidth]{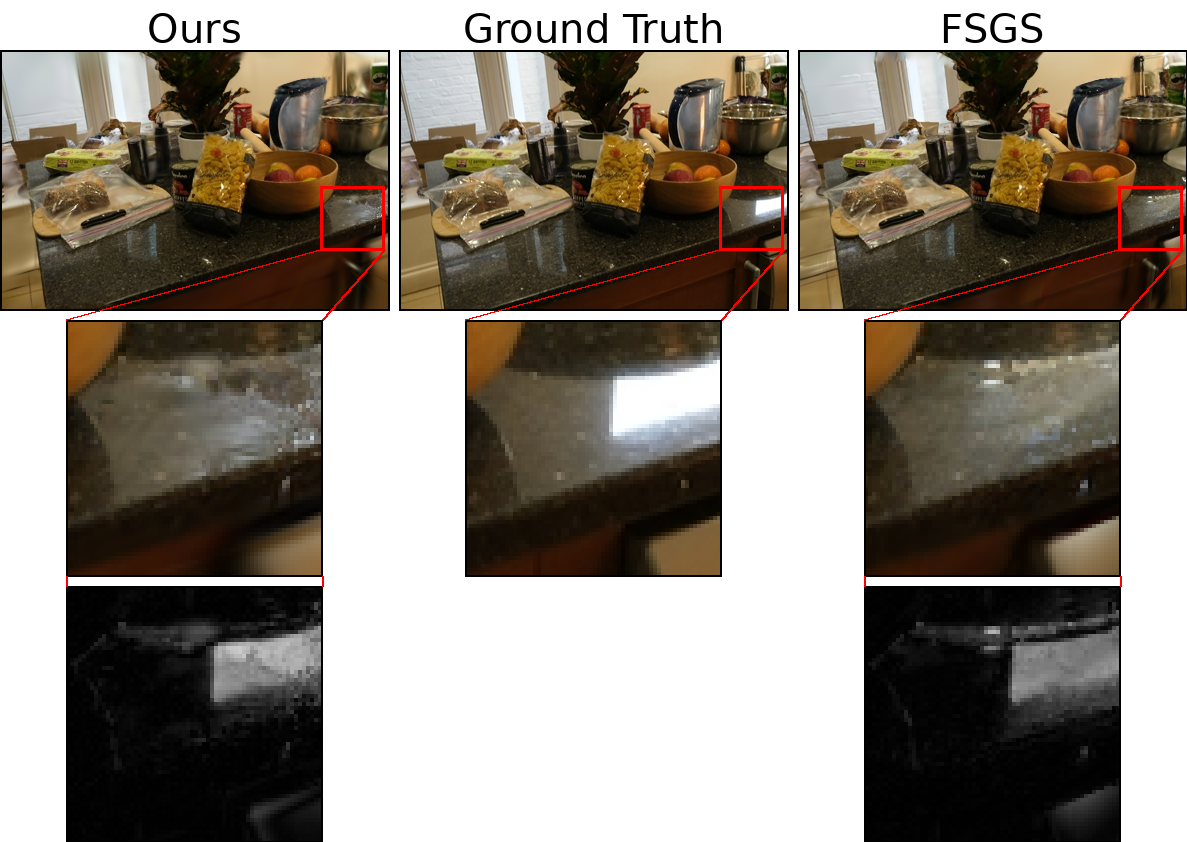}
        \caption{Mip-NeRF 360 Counter (Poor Case)}
        \label{fig:mipnerf_counter_poor}
    \end{subfigure}
    \caption{Poor-performance (failure case) results. These examples illustrate the framework's limitations. When faced with highly ambiguous geometry or poor initial priors from SfM, our method can produce artifacts such as floaters or geometric inaccuracies.}
    \label{fig:qualitative_poor}
\end{figure*}

\subsubsection{Best-Case Performance}
Figure \ref{fig:qualitative_best} illustrates our framework's performance in ideal scenarios. In scenes with good initialization from SfM and distinct geometric features, such as the `flower` scene from LLFF and the `garden` scene from Mip-NeRF 360, our framework excels. The error-driven densification correctly identifies regions of high reconstruction error, such as object silhouettes and textured surfaces, allocating new primitives efficiently. The conservative pruning schedule allows these new Gaussians to optimize their positions, shapes, and colors, contributing to a sharp, detailed final render. The resulting model is highly compact, containing only the necessary primitives to represent the scene, which is a direct consequence of our efficiency-focused optimization cycle. The low error maps in these cases confirm that high visual fidelity is achieved with a fraction of the primitive count compared to FSGS.

\subsubsection{Marginal-Case Performance}
Figure \ref{fig:qualitative_marginal} shows more challenging scenarios. In scenes characterized by very fine, repetitive textures (`room`) or complex, semi-transparent surfaces (`counter`), our framework produces results that are quantitatively strong but may lack the high-frequency detail of more complex models like FSGS. This outcome is a direct trade-off made by our optimization strategy. The opacity-gradient-driven densification prioritizes larger areas of error, and with a strict primitive budget, it may not allocate enough Gaussians to capture every minute detail. The error maps highlight these subtle differences, often localized in areas of complex material properties. This demonstrates a core principle of our framework: we accept a modest and often imperceptible decrease in fidelity on challenging textures in exchange for a significant reduction in model complexity and rendering time.

\subsubsection{Failure Cases and Limitations}
Figure \ref{fig:qualitative_poor} presents failure cases, which are critical for understanding the limitations of our framework. Our framework's performance is fundamentally dependent on the quality of the initial geometric priors from SfM and the monocular depth estimator. In scenes with severe challenges, such as the ambiguous geometry in the `leaves` scene or reflective surfaces in the `counter` scene, the initial point cloud can be sparse and inaccurate. In these situations, our error-driven densification can be misled. If the depth priors are incorrect, the opacity gradients may point to empty space, causing the framework to allocate primitives ("floaters") that do not correspond to real-world geometry, as shown in the error maps. While our conservative pruning mitigates this to some extent, it cannot fully compensate for a fundamentally flawed geometric initialization. This demonstrates that while our optimization is more efficient, it is not a complete solution for the ill-posed nature of few-shot reconstruction and remains sensitive to the quality of its inputs.

\section{Conclusion, Limitation and Future Work}
This paper addressed the research question of how to reformulate the core 3DGS algorithm for geometric efficiency in sparse-view scenarios. We hypothesized that a synergistic redesign of the Adaptive Density Control, pairing an error-driven densifier with a conservative pruner, could prevent model bloat and produce more compact representations. Our results validate this hypothesis. The key contribution is the empirical demonstration of the critical interplay between densification and pruning; our proposed ADC creates an optimization cycle that is fundamentally more efficient in data-sparse regimes. By integrating this improved core algorithm with a standard depth-correlation loss, our final framework establishes a new state-of-the-art on the efficiency-quality Pareto frontier, achieving a model compactness reduction of over 40\% on the LLFF dataset and approximately 70\% on the Mip-NeRF 360 dataset, with only a modest trade-off in image quality. This work provides a more principled and efficient foundation for few-shot novel view synthesis.

\textbf{Limitations.} The framework's most dramatic efficiency gains are observed in extremely sparse settings like the 3-view LLFF dataset. While it remains competitive in less sparse scenarios, the compactness benefits may be less pronounced. Furthermore, the framework relies on geometric priors from an external monocular depth estimator, which is a common dependency but one that prevents a fully end-to-end solution. Finally, the optimal ADC hyperparameters (e.g., pruning delay and thresholds) were determined empirically and may require tuning for different datasets or application requirements.

\textbf{Future Work.} The findings of this paper open several promising avenues for future research. A primary direction is the integration of the error-driven densification \textit{trigger} with alternative densification \textit{mechanics}, such as the proximity-based unpooling from FSGS. This could potentially combine the benefits of intelligent placement with geometrically aware growth. Another key area is the development of a dynamic, or even learned, pruning schedule that can adapt to the scene complexity and training progress, removing the need for fixed hyperparameters. Finally, exploring methods to reduce or eliminate the reliance on external geometric priors would be a significant step towards a more robust and self-contained system for few-shot view synthesis.

\end{document}